\documentclass[letterpaper, 10 pt, conference]{ieeeconf}  % Comment this line out if you need a4paper

\IEEEoverridecommandlockouts                              % This command is only needed if
\usepackage{graphics} % for pdf, bitmapped graphics files
\usepackage{epsfig} % for postscript graphics files
\usepackage{epstopdf}
\usepackage{amsmath} % assumes amsmath package installed
\usepackage{amssymb}  % assumes amsmath package installed
\usepackage{booktabs}
\usepackage{multirow} % for tables
\newcommand{\RNum}[1]{\uppercase\expandafter{\romannumeral #1\relax}}

\makeatletter
\let\NAT@parse\undefined
\makeatother
\usepackage{hyperref}
\hypersetup{hypertex=true,
            colorlinks=true,
            linkcolor=black,
            anchorcolor=black,
            citecolor=black}

\title{\LARGE \bf
GP-SLAM+: real-time 3D lidar SLAM based on improved regionalized Gaussian process map reconstruction
}

\author{Jianyuan Ruan\textsuperscript{1}, Bo Li\textsuperscript{2,1}, Yinqiang Wang\textsuperscript{1}, and Zhou Fang\textsuperscript{1}% <-this % stops a space
\thanks{1, School of Aeronautics and Astronautics, Zhejiang University, China.}%
\thanks{2, School of automation and electrical engineering, Zhejiang university of science \& technology, China.}%
\thanks{{\tt\small \{ruanjy, 11224012, wangyinqiang, zfang\}@zju.edu.cn }}
\thanks{The authors would like to thank Prof. Yu Zhang for devices.}
}

\begin{document}

\maketitle
\thispagestyle{empty}
\pagestyle{empty}

%%%%%%%%%%%%%%%%%%%%%%%%%%%%%%%%%%%%%%%%%%%%%%%%%%%%%%%%%%%%%%%%%%%%%%%%%%%%%%%%
\begin{abstract}

This paper presents a 3D lidar SLAM system based on improved regionalized Gaussian process (GP) map reconstruction to provide both low-drift state estimation and mapping in real-time for robotics applications. We utilize spatial GP regression to model the environment. This tool enables us to recover surfaces including those in sparsely scanned areas and obtain uniform samples with uncertainty. Those properties facilitate robust data association and map updating in our scan-to-map registration scheme, especially when working with sparse range data. Compared with previous GP-SLAM, this work overcomes the prohibitive computational complexity of GP and redesigns the registration strategy to meet the accuracy requirements in 3D scenarios. For large-scale tasks, a two-thread framework is employed to suppress the drift further. Aerial and ground-based experiments demonstrate that our method allows robust odometry and precise mapping in real-time. It also outperforms the state-of-the-art lidar SLAM systems in our tests with light-weight sensors.

\end{abstract}

%%%%%%%%%%%%%%%%%%%%%%%%%%%%%%%%%%%%%%%%%%%%%%%%%%%%%%%%%%%%%%%%%%%%%%%%%%%%%%%%
\section{Introduction}

Simultaneous localization and mapping (SLAM) is one of the essential functions for autonomous robots. Its primary tasks are state estimation and map building. State estimation aims at finding the transformation that best aligns consecutive sensor data, in which a data association process is required. Map building involves representing the environment using a specific type of the model and accumulating information. The chosen model of the environment is fundamental for data association, and thus it impacts the accuracy and efficiency of the whole system. 

Specifically, in lidar SLAM problem, point set registration \cite{pomerleau2015review} is needed for state estimation. For mobile robot with light-weight sensors and limited computational resource, it is challenging to achieve accurate data association efficiently due to the sensor mechanism and the motion of the vehicles. For instance, the spinning 2D lidar \cite{droeschel2016multilayered} provides low-resolution point cloud in low scan rate. The single-axis 3D lidar, for example, VLP-16 used in \cite{shan2018lego}, also produces data with low vertical resolution, which means the range data still aggregate in several channels due to the sweeping mechanism. Regarding the map building, it is critical to avoid the dimension explosion of the map state vector when accumulating the data into it. Consequently, when facing large amount non-uniform and sparse data, it is still worth pursing more reliable registration method suitable for both state estimation and map building.

\begin{figure}[t]
	\centering
	\includegraphics[scale = 0.95]{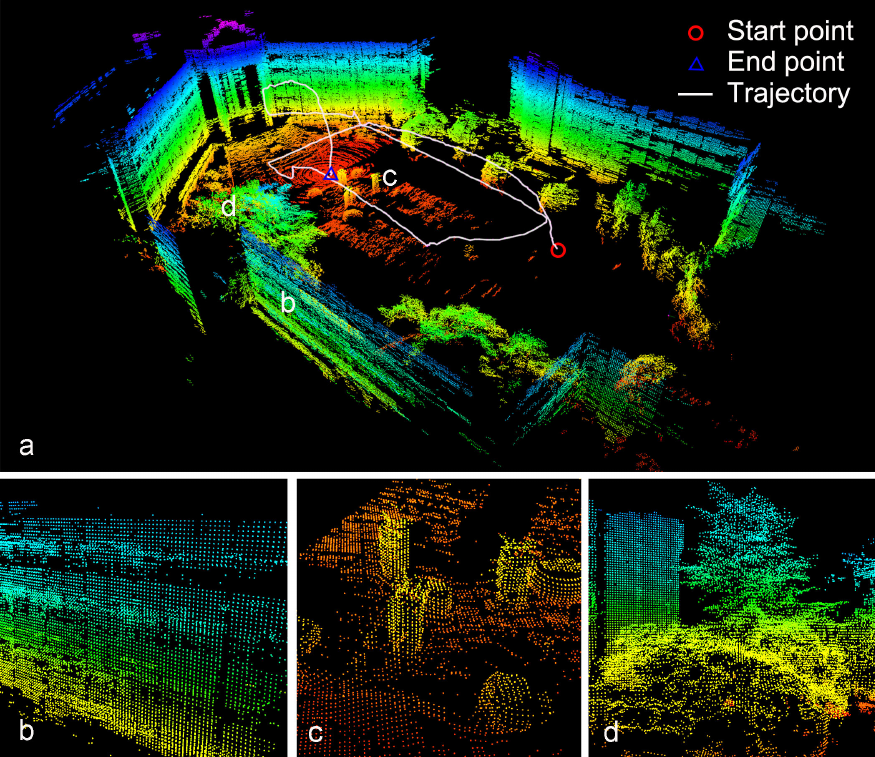}
	\caption{Map of a plaza reconstructed by GP. We show those points whose uncertainty are below certain threshold. They are uniformly distributed and colored according to height. (a) is a perspective view of the map produced by the core workflow. The white curve indicates the trajectory of the MAV. (b-d) are the detailed views of several typical objects in environment including a facade of building (b), sculptures (c), and unstructured trees (d).}
	\label{fig:1}
	%\setlength{\belowcaptionskip}{5cm}
	%\vspace{-0.5cm}
\end{figure}

In this work, a 3D lidar-based SLAM approach, named GP-SLAM+, is designed to address those challenges above. We use regionalized GP map reconstruction to model the environment from range data, which serves as the fundamental of our approach. After this, evenly distributed samples are drawn from the model and fed into a scan-to-map registration scheme to compute the rigid transformation. Map is built incrementally by fusing the information in current frames into it. One of the mapping results can be seen in Fig. \ref{fig:1}. This GP-SLAM workflow was proposed in our previous work \cite{li2020gp} in 2D situation. We also investigated the registration between dense 3D point clouds \cite{Li2020}. However, moving to 3D space, the structure is more complicated and thus can not represented by a function easily. Also, the cubic complexity of the GP becomes prohibitive. This work overcomes those barriers. Firstly, we use a principled down-sample method to accelerate the training of GP. The registration, including the data association, is redesigned based on a maximum likelihood estimation (MLE) probabilistic scheme. We also design a two-thread framework to further enhance the mapping quality and the fidelity of pose estimation in large-scale tasks. We implemented experiments with light-weight sensors to thoroughly evaluate the core workflow and full system.

\section{Related Work}

A wide range of existing literature devoted to build lidar-based SLAM systems. Many of them are based on the iterative closest point (ICP) method \cite{besl1992method} or its variants \cite{rusinkiewicz2001efficient}. Classical ICP may fall into local minima caused by the sparsity of range data. Consequently, it is more recommended to identify more stable features to capture the environment structure. Geometric features, such as lines and planes, can be extracted easily and are used widely. These features are incorporated into a probabilistic framework by Generalize ICP (GICP) \cite{segal2009generalized}. Lidar Odometry and Mapping (LOAM) \cite{zhang2014loam} is one of the state-of-the-art systems that extract such features. Then LeGO-LOAM \cite{shan2018lego} avoids features extracted from noisy areas like vegetation. Another option is to study the properties of point cloud within sub-sections. Normal distribution transform (NDT) -based methods  \cite{magnusson2007scan}\cite{hong2017probabilistic} and surfel-based method \cite{droeschel2016multilayered}\cite{bosse2009continuous} fall into this category. It is also notable that \cite{deschaud2018imls} constructs a implicit surface for precise registration offline. By contrast, our GP-based mapping use several GPs in sub-domains to express the local surfaces. It reduces the loss of information caused by feature extraction. In the other hand, this model can fully recover the structure well with a fixed grid size compared with those multi-resolution grid-based parametric model \cite{droeschel2016multilayered}\cite{hong2017probabilistic}\cite{bosse2009continuous}.

GP-based mapping appeals notice in robotics society in recent years as it is a continuous representation and can make inference with uncertainty in un-explored regions. Several works use GP as a regressor to obtain continuous occupancy grid map \cite{o2012gaussian}, or use it to model and interpolate the strength of ambient magnetic field \cite{solin2018modeling} for indoor localization, while we use spatial GP to recover the local surfaces directly. Some other works use spatial GP on terrain modeling \cite{plagemann2008learning} or surface reconstruction \cite{lee2019online}. The functional relationships in our method share certain similarity with those works. However, they mainly focus on mapping problem and do not include real-time state estimation in a SLAM system. Also, few methods complete large-scale 3D mapping online.

With the aforementioned representations of the environment, the amount of components in registration is reduced, and the efficiency is enhanced. However, most feature-based methods still suffer from the time-consuming matching process and usually use data structures like Kd-tree \cite{bentley1975multidimensional} to accelerate it. Computational complexity is also the main occlusion that prevents GP from wider robotics applications. Domain decomposition \cite{park2011domainDec} and local regression \cite{shen2006fast} with Kd-tree are two techniques that improve the efficiency of GP. Our method adopts both techniques. However, in this work, by utilizing the evenly distributed property of the samples from GP map reconstruction, the matching can be finished directly and the Kd-tree is also avoided.

\section{GP Map Reconstruction}

We use spatial GP to reconstruct local surfaces from noisy range observations. To get more feasible access to data association and map updating, we extract discrete samples from recovered surfaces. This process is named as regionalized GP map construction. In other words, it can be considered as certain kind of surface interpolation. This process contains two components, regionalization and reconstruction.

\begin{figure}
	\centering
	\includegraphics[scale = 0.60]{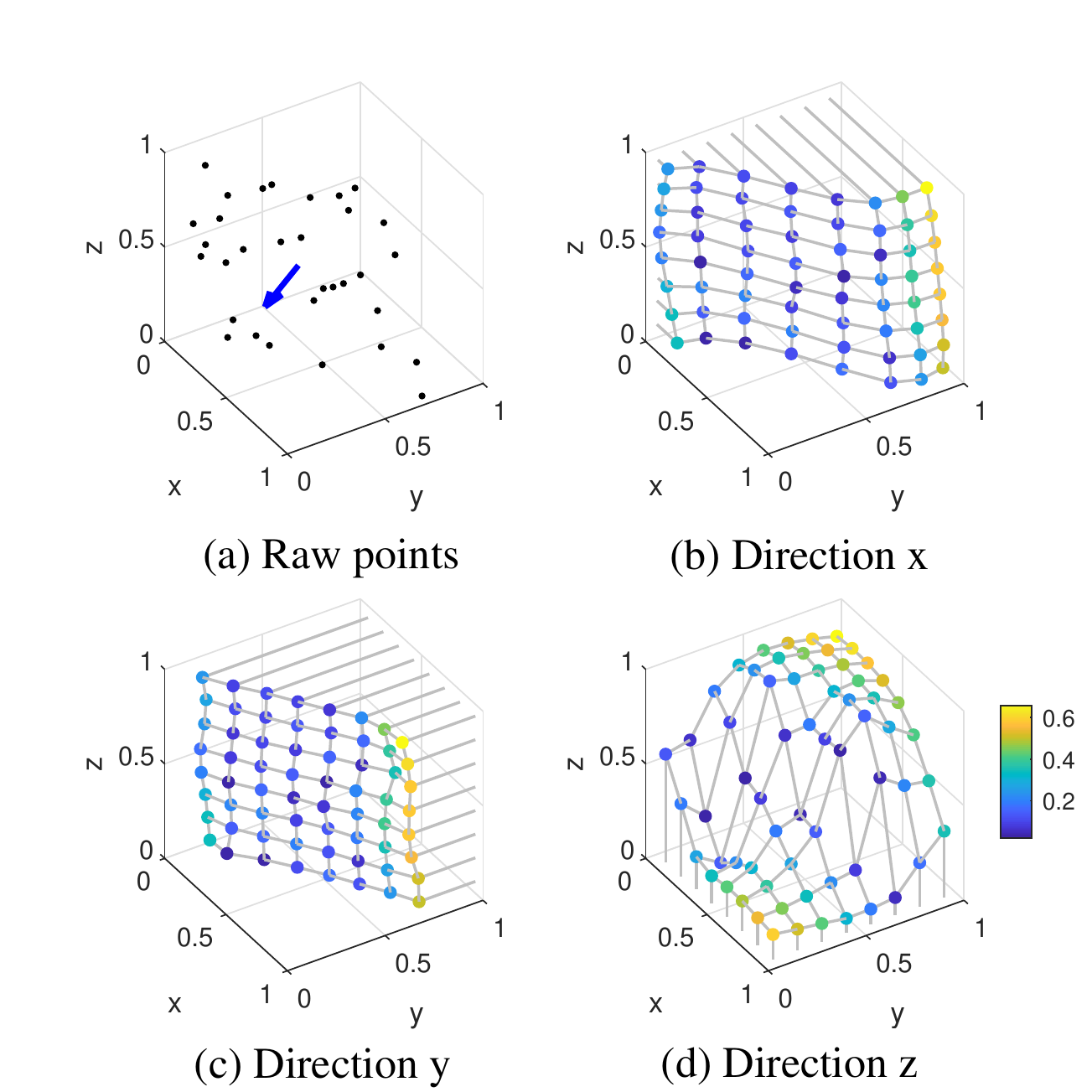}
	\caption{Illustration of the GP map reconstruction process in a cell. (a) Sparse raw points indicated by black points. The blue arrow refers to the normal from PCA. This local surface is approximately perpendicular to the $oxy$ plane. (b-d) show the results of GP map reconstruction in each direction. Samples are colored according to its variance. The direction $z$ is determined to be noisy direction as shown in (d) and will be omitted.}
	\label{fig:2}
	%\vspace{-0.6cm}
\end{figure}

\subsection{Regionalization}

Initially, to establish different function relationships locally, we divide the whole domain into several evenly distributed cubic cells in the word coordinate system $\{W\}$. This decomposition also accelerates the reconstruction process \cite{park2011domainDec}. The side length of each cell is $a$. The subset of the raw points $S_{t}^{W}$ located in the $k$th cell is denoted by $S_{t, k}^{W}=\left\{p_{k, i}, i=1, \dots, n_{k}\right\}$.

Then we need to determine the function relationship between the coordinates as $x=f(y,z)$, or $y=f(x,z)$, or $z=f(x,y)$. Considering one function can only express a 2.5D surface, in case of complex 3D structure, generally we assume three functions exist in a cell. Each function provides corresponding constrains along its direction, which will be detailed in the Section \RNum{4}. Accordingly, if the surface in a cell is perpendicular to one coordinate plane, as it only provides constrains along its normal, we can omit the corresponding function in this cell. This situation is judged based on the principled component analysis (PCA). Fig. \ref{fig:2} illustrates a example that when a set of raw data are drawn from a vertical wall, the function whose direction is $z$ will be neglected as the wall cannot provide vertical constrains. 

\subsection{Reconstruction}

After the regionalization, we conduct GP map reconstruction in each nonempty cell. Lidar measurements are noisy samples of the environment. The noise model of it can be derived from manufacture data as in \cite{hong2017probabilistic}. Here, we simply assume that each lidar point follows an independent normal distribution with a isotropic variance $\sigma^{2}$. In this case, GP regression, which can produce the best linear unbiased prediction \cite{park2011domainDec}, is used.
 
The GP regression problem is detailed as follows based on \cite{rasmussen2003gaussian}. Given $n_{k}$ training points as $D=\left\{\left(f_{i}, {l}_{i}\right), i=1, \dots n_{k}\right\}$, the relationship between the observations $f_{i} \in \mathbb{R}$ in the training locations $l_{i} \in \mathbb{R}^{2}$ is expressed as $f_{i}=f\left({l}_{i}\right)+\varepsilon_{i}, i=1, \dots, {n}_{k}$, where $\varepsilon_{i}$ is the noise term following the distribution of $\varepsilon_{i} \sim \mathcal{N}\left(0, \sigma^{2}\right)$. The goal is to achieve the distribution of $n_{{test}}$ predictions $\boldsymbol{f}_{*}$ in the preset test locations $\boldsymbol{l}_{*}=\left[l_{* 1}, l_{* 2}, \ldots, l_{* n_{ {test}}}\right]^{T}$ denoted by $f_{* j} =f\left( l_{* j}\right)+\varepsilon_{* j}, j=1, \ldots, {n}_{ {test}}$. Defining $\boldsymbol{f}=\left[f_{1}, f_{2}, \ldots, f_{{n}_{k}}\right]^{T}$, $\boldsymbol{l}=\left[l_{1}, l_{2}, \ldots, l_{{n}_{k}}\right]^{T}$ the predictive distribution $\boldsymbol{f}_{*}$ of given $\boldsymbol{f}$ will be
\begin{equation}
\begin{array}{r}
P\left(\boldsymbol{f}_{*} | \boldsymbol{f}\right)=\mathcal{N}\left(k_{\boldsymbol{l *}}^{T}\left(\sigma^{2} I+K_{\boldsymbol{l l}}\right)^{-1} \boldsymbol{f}, k_{\boldsymbol{* *}}-\right. \\
\left.k_{\boldsymbol{l *}}^{T}\left(\sigma^{2} I+K_{\boldsymbol{l l}}\right)^{-1} k_{\boldsymbol{l *}}\right)
\end{array}
\end{equation}
in which the mean value $k_{\boldsymbol{l *}}^{T}\left(\sigma^{2} I+{K}_{\boldsymbol{ll}}\right)^{-1} \boldsymbol{f}$ is taken as the point prediction of $\boldsymbol{f}_{*}$ at test locations $\boldsymbol{l}_{*}$. Its variance is estimated by $k_{\boldsymbol{**}}-k_{\boldsymbol{l *}}^{T}\left(\sigma^{2} I+{K}_{\boldsymbol{l} \boldsymbol{l}}\right)^{-1} k_{\boldsymbol{l *}}$. Here, $k_{\boldsymbol{**}}=k\left(\boldsymbol{l}_{*}, \boldsymbol{l}_{*}\right)$, $k_{\boldsymbol{l*}}={k\left(\boldsymbol{l}, \boldsymbol{l}_{*}\right)}^{T}$ and $\boldsymbol{K}_{\boldsymbol{l} \boldsymbol{l}}$ is an $n_{k} \times n_{k}$ matrix, $\boldsymbol{K}_{\boldsymbol{l} \boldsymbol{l}}(i, j)=k(i, j)$, $ k(., .)$ represent the kernel function. In this work, we choose the commonly used exponential covariance function, $k\left(l_{i}, l_{j}\right)=\exp \left(-\kappa\left|l_{i}-l_{j}\right|\right)$, with a preset length-scale parameter $\kappa$.

In this context, the training points are the raw points $S_{t, k}^{W}$ in a cell. The coordinate used as observation is named direction, and the other two serve as a training location. The $n_{{test}}$ test locations are evenly set, and the interval between them is $r$. Recalling the side length of the cell is $a$, we set $a$ as integral times of $r$, which means $n_{ {test}}=({a} / {r})^{2}$. Those predictions with variance are samples drawn from the implicit surfaces, and each set of samples are named as layer. As shown in Fig. \ref{fig:2}, those predictions which are remote from raw data are more unreliable. We use these samples as the reconstruction result.

After the reconstruction, there are 0$\sim$3 layers in one cell. The cells is stored in a hashing table data structure. A sample is represented by $p_{i}=\left(f_{i}, l_{i}\right)$, where $f_{i} \sim \mathcal{N}\left({u}_{i}, \sigma_{i}^{2}\right)$ and the test location serve as index. By this way, a sample can be queried directly.

\subsection{Acceleration of the Reconstruction}

 Besides the domain decomposition, we further accelerate the training process of GP through the concept of local regression. The central idea is that a prediction is mostly influenced by those observations whose training locations are closer to the test location of that prediction. Thus, the training process can be accelerated by principled down-sample of the training points without much precision loss \cite{shen2006fast}. Accordingly, we retain the raw data but only use all the closest points of each test location in GP map reconstruction each time. As a result, the amount of the filtered training points is reduced significantly.

\begin{figure}
  \centering
  \includegraphics[scale = 0.9]{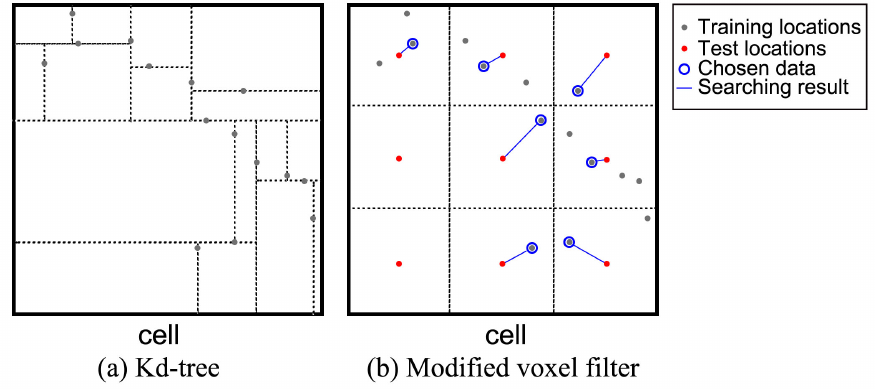}
  \caption{Illustration of the principled filtering process. (a) Kd-tree divides the searching domain according to data, (b) Our modified 2D voxel filter divides it into smaller grid indicated by the dash lines, and the centers of them are the test locations (red points). A raw point will be kept if its training location (gray point) is the closest one to the test location in a smaller grid, the blue line indicates this relationship, the filtering result is highlighted by the blue circles. }
  \label{fig:3}
%\vspace{-0.4cm}
\end{figure}

One approach to complete this filtering process is utilizing the Kd-tree. However, the initializing cost of this data structure is $O({n{log}n})$ with $n$ inputs, and the average searching cost is $O({log}n)$ \cite{bentley1975multidimensional}. Although Kd-tree is faster than the brute-force searching, it is still time-consuming especially when the amount of points is large. As the searching targets in our application are evenly distributed, we use a modified 2D voxel filter to approximate this process. As shown in Fig. \ref{fig:3}, in a cell, the training locations spread over a 2D domain, which is divided into smaller grids whose center are the test locations. The original voxel filter calculates the mean of all raw data in each smaller grid. The modification is that we keep a point if it is the closest one to the test location among all points in the same smaller grid. By this way. The filtering process can be finished with linear complexity cost.

\section{State Estimation}

Follow the reconstruction, the current frame is aligned to the map. Using the map as the reference frame, we can suppress the pose drift. The GP map reconstruction and scan registration processes are conducted iteratively till it converges to provide the state estimation. This scan-to-map registration process includes two main steps, matching and alignment. The matching step establishes the correspondences between the two frames, and the alignment step targets on computing a transformation between the matched pairs.

\subsection{Matching}

The correspondences will be established between two samples coming from the current frame $P_{t}^{W}$ and the reference frame $Q_{t-1}^{W}$ respectively when they satisfy the following conditions: 1) two samples are located in the same or adjacent cell; 2) two samples share the same prediction direction and test location; 3) both variances of the samples are below threshold $\sigma_{thr}^{2}$. We illustrate this process in Fig. \ref{fig:4} in 2D space for simplicity. Pair-1 is a qualified correspondence while Pair-2 is invalid as the variance of one sample is too large. Pair-3 is established between two samples as they share another direction and are located in adjacent cells. In Pair-1, there are several samples satisfy aforementioned conditions. In this case the closer one is chosen. Paired samples are expressed by $\{p_{i}, q_{i}\}$, where $p_{i}=\left(f_{pi}, l_{pi}\right)$,
 $f_{p i} \sim \mathcal{N}\left({u}_{p i}, \sigma_{p i}^{2}\right)$ and $q_{i}=\left(f_{qi}, l_{qi}\right)$,
 $f_{q i} \sim \mathcal{N}\left({u}_{q i}, \sigma_{q i}^{2}\right)$.

\subsection{Alignment}

Based on the idea that layers only offer observability in their directions, we design the error metric as the distance between prediction of samples. Firstly, Given $e_{x}=[1,0,0]$, the coordinate $x_{i}$ of a 3D point $p_{i}={[x_{i},y_{i},z_{i}]}^{T}$ can be computed by $x_{i}=e_{x} \cdot p_{i}$, and the other two directions are the same. For the sake of brevity, we define an operator $\left(  \cdot \right)_{\cdot} {\circ}$ to express such operation of obtaining the coordinate that corresponds to the direction $\circ$ in the following context.  Similar to the MLE approach expressed in \cite{segal2009generalized}, with $n_{cur}$ paired samples $\{p_{i}, q_{i}\}, i=1, \dots, n_{c u r}$ in all layers, for a transformation $\boldmath{T}$, we define the 1-dimensional distribution of an observation in certain test location as ${d}_{i}\sim\mathcal{N}\left({\left(T {p}_{i}\right)}_{\cdot}\circ-q_{i\cdot}\circ, \sigma_{p i}^{2}+\sigma_{q i}^{2}\right)$. Then the relative transformation is compute by
\begin{equation}
{T}=\underset{T}{{argmax}} \prod_{i}\left({P}\left({d}_{i}\right)\right)\\
=\underset{T}{{argmax}} \sum_{i} \log \left({P}\left({d}_{i}\right)\right)
\end{equation}
The above objective function can be simplified to
\begin{equation}
\begin{aligned}
{T}&=\underset{T}{{argmin}} \sum_{i}\left({{d}_{i}}^{T}\left(\sigma_{p i}^{2}+\sigma_{q i}^{2}\right)^{-1} {d}_{i}\right) \\
&=\underset{T}{{argmin}} \sum_{i} \frac{\left\|\left(T {p}_{i}\right)_{\cdot}{\circ}-q_{i \cdot} \circ \right\|^{2}}{\sigma_{p i}^{2}+\sigma_{q i}^{2}}
\end{aligned}
\end{equation}
where the variance can be seen as weight. This optimization problem is solved by the non-linear solvers Ceres \cite{ceres-solver}.

\begin{figure}
	\centering
	\includegraphics[scale = 0.22]{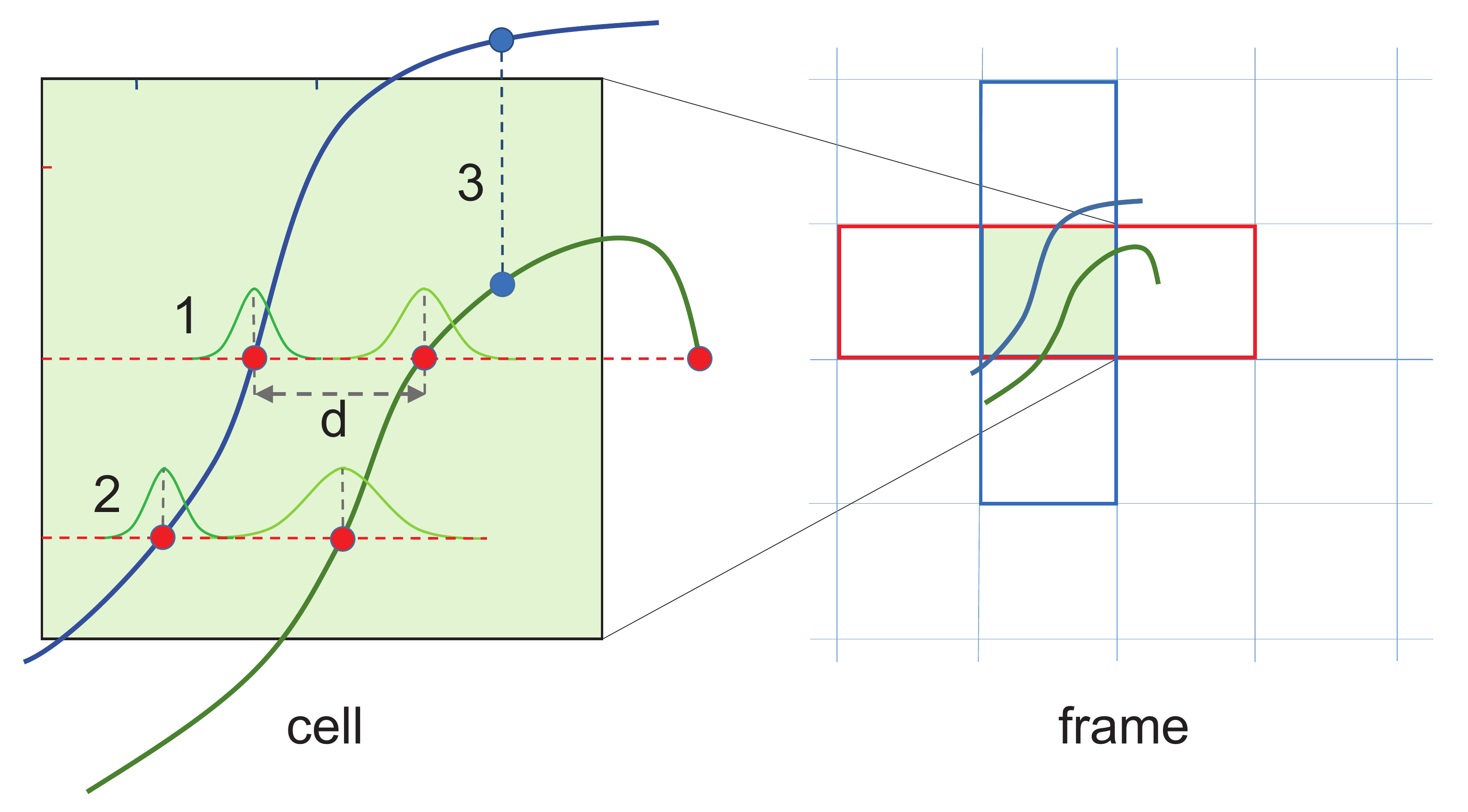}
	\caption{Illustration of the matching and error metric in 2D situation for simplicity. In the cell view, points indicate samples drawn from surfaces with uncertainty along its direction, and the dash lines refer to the identical test locations between established correspondences. In the frame view, searching happen in identical and adjacent several cells represented by the colored rectangles. The color of points, dash lines and rectangles indicate two different directions.}
	\label{fig:4}
	%\vspace{-0.4cm}
\end{figure}

Compared with our previous work, the registration has been redesigned in several aspects. In the previous work, the correspondences are established using only one layer within each cell. It treats all the reconstructed samples as 3D points, and use the 3D Euclidean distance as the error metric. The problem is solved by singular value decomposition (SVD). In contrast, we use several layers to model complex structure and extended the correspondences searching area to avoid information loss in borders (In the previous work, the pair-3 in Fig. \ref{fig:4} is omitted). The error metric is also changed so that it will drags the surfaces, rater than points as in previous work, closer. This leads to faster convergence as it avoids inducing the test locations into the objective function.

\subsection{Demonstration of Registration}

\begin{figure}
  \centering
  \includegraphics[scale = 0.95]{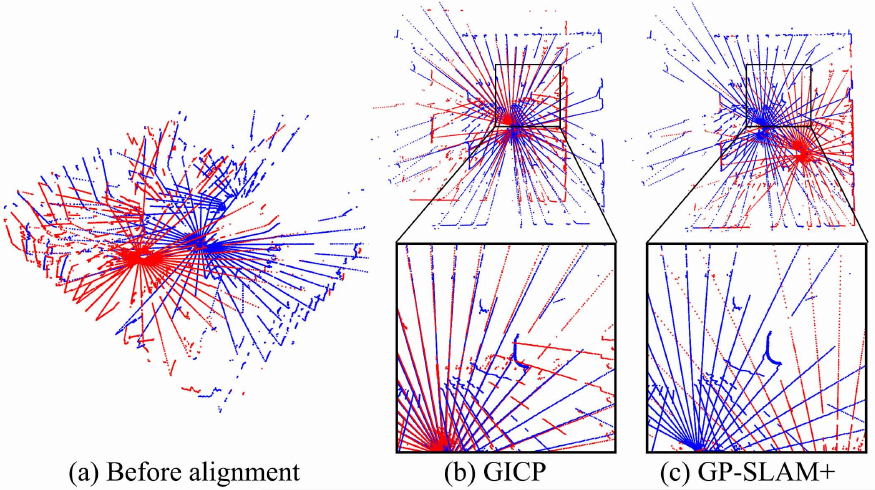}
  \caption{Registration test of sparse point cloud from a spinning 2D lidar. Red and blue points indicate the two frames respectively. (a) Two frames of data before alignment. (b) Result of GICP. Although the points are drawn closer, the walls and columns are misaligned. (c) Registration of GP-SLAM+ outputs correct result.}
  \label{fig:5}
\end{figure}

We select two sets of typical range data to demonstrate the advantages of our registration upon ICP and previous method. In the first test, we use two frames of range data measured with a spinning 2D lidar from experiment A in Section \RNum{7}. As shown in Fig. \ref{fig:5}(a), the point cloud provided by this kind of sensor is rather sparse and uneven. Here, the Generalized ICP (GICP) is selected as the benchmark. The results in Fig. \ref{fig:5}(b)\&(c) shows that GICP falls into wrong local minima, while our method align structure well.

\begin{figure}
  \centering
  \includegraphics[scale = 0.45]{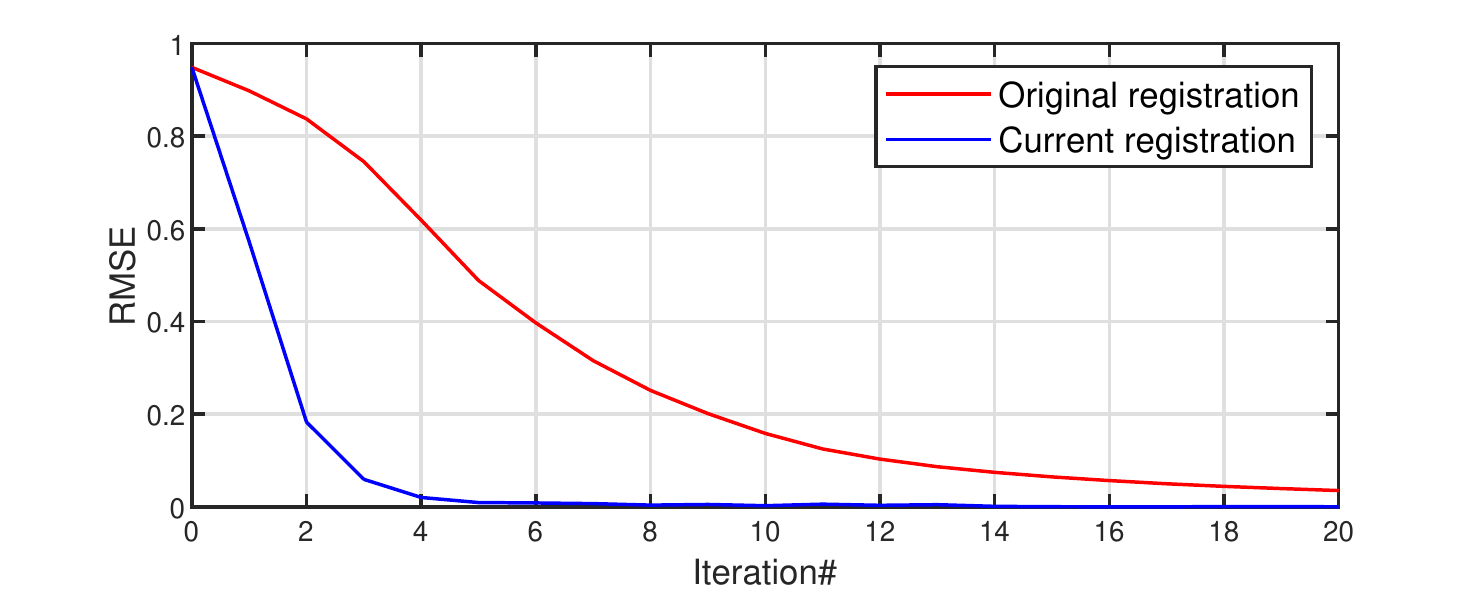}
  \caption{Comparison of converge speed between original and current registration methods. The RMSE between the closest point pairs from target frame and the source frame after each iteration is used as the convergence criteria. }
  \label{fig:6}
  %\vspace{-0.6cm}
\end{figure}

Secondly we check the impact of our modification on registration strategy. We choose a frame of range data collected by Velodyne VLP-16 from experiment C as the target frame and set an initial transformation error (1m in translation and 5 degree in rotation around the $z$-axis) on it to form the source frame. Then these two frames are aligned using the original and current registration methods. The root mean square error (RMSE) of the distances between the closest points from two frames are used as the convergence criteria. As shown in Fig. \ref{fig:6}, the result indicates that the registration part of our method is significantly improved.

\section{Map Building}

The map represents the accumulation of historical information. We build it incrementally making use of the uncertainty and convenient data association approach again. The map is initialized by the first frame after map reconstruction. The following current frame $P_{t}^{W}$ is fused into $Q_{t-1}^{W}$ to form the updated map $Q_{t}^{W}$.

In detail, there are three different cases: The newly explored cells or newly built layers are added to  $Q_{t-1}^{W}$ directly; In those overlapping cells, two layer with the same direction are fused by a recursive least square method. For two samples $\{ p_{cur}, q_{map}\}$ sharing the same test location from these two layers, we obtain the updated sample $q_{upd}$ by:

\begin{equation}
\begin{aligned}
\sigma_{upt}^{2}=\frac{\sigma_{map}^{2} \sigma_{cur}^{2}}{\sigma_{map}^{2}+\sigma_{cur}^{2}},
\end{aligned}
\end{equation}

\begin{equation}
\begin{aligned}
f_{upt}=\frac{\sigma_{{map}}^{2} f_{cur^{+}} \sigma_{cur}^{2} f_{{map}}}{\sigma_{{map}}^{2}+\sigma_{cur}^{2}},
\end{aligned}
\end{equation}
where $f_{upt}$ and $\sigma_{upt}^{2}$ refer to the prediction and variance of the updated sample; In the last case, two overlapping cells in both frames contain only raw data, implying that these raw data was too sparse, we accumulate the data and conduct the reconstruction.

Since we only update the predictions of the samples, and their test locations are fixed, the dimension of the map state in each voxel cell is prevented from exceeding the number of test locations as the SLAM process unfolded.

\section{Two-Thread Framework}
\begin{figure}
  \centering
  \includegraphics[scale = 0.37]{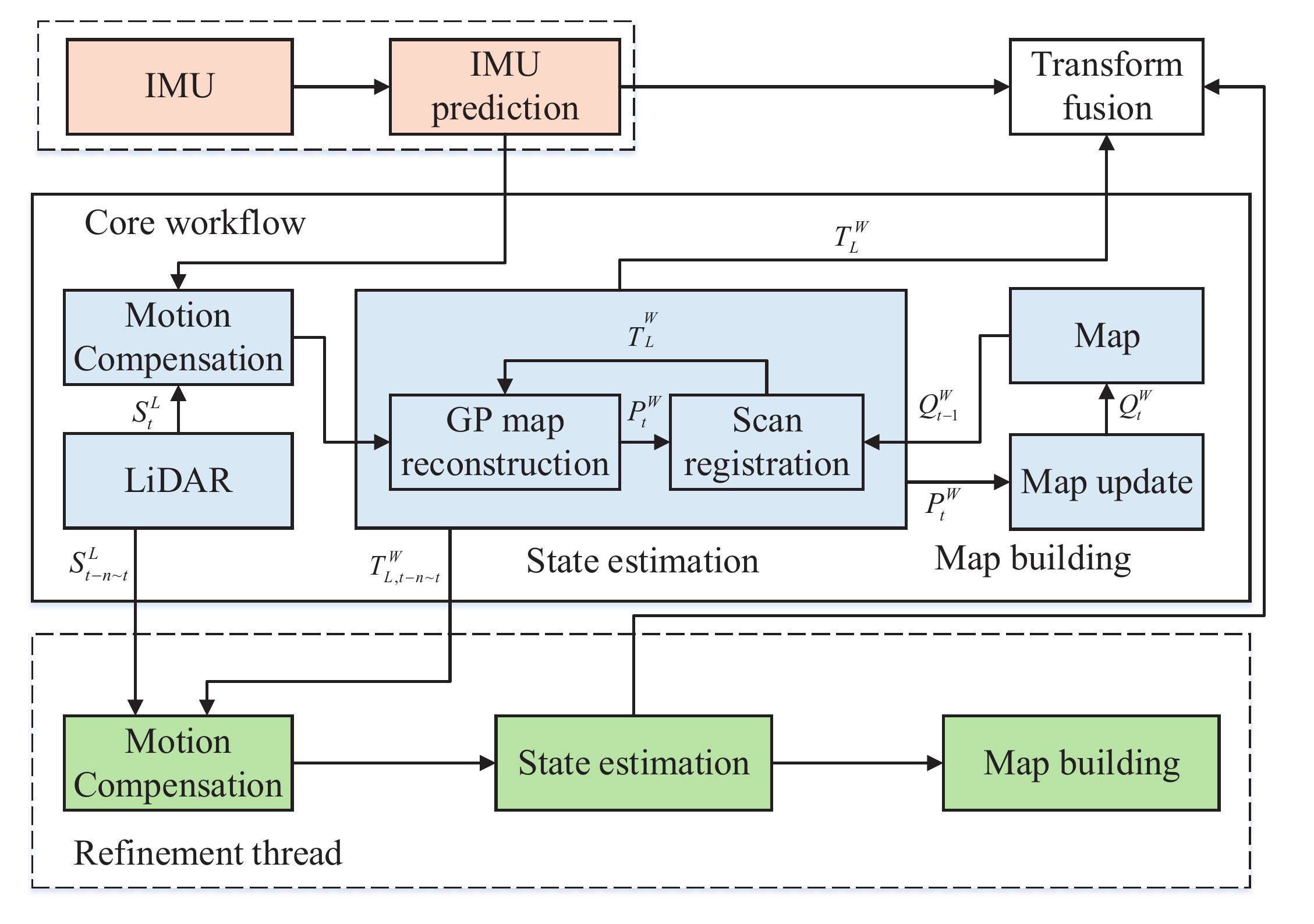}
  \caption{Architecture of the full system. The central block is the core workflow. IMU or the two-thread framework is optional as the core workflow can finish odometry and mapping independently. }
  \label{fig:7}
  %\vspace{-0.6cm}
\end{figure}

Although the core workflow can complete low-drift odometry and dense mapping independently, when IMU or multi-core hardware is available, it can be extended to the full system. The prediction of IMU can compensate motion distortion and provide initial guess $\hat{T}_{L, t}^{W}$ for scan registration. Subsequently, the result of scan registration is fed back to rectify the bias. The two-thread architecture can further enhance the mapping quality and decrease the drift, especially in large-scale scenarios. The full architecture is shown in Fig. \ref{fig:7}.
This framework is inspired by \cite{zhang2014loam}. However, in contrast to it, both threads in our system use the same scan-to-map strategy described in Section \RNum{3}-\RNum{5}, so the state estimation produced by our core workflow perform higher fidelity than the odometry thread in \cite{zhang2014loam}.

More precisely, after the core workflow has processed with several sequential frames of point clouds, those aligned points and the relative transformation are sent to the refinement thread. As the aggregated point cloud is denser, the variances of the samples are smaller as the GP map reconstruction can reveal the real structure of the environment better, and there are more valid constrains. As the refinement thread operates in a lower frequency (2 Hz in our implementation), to obtain more accurate odometry results, the registration module in this thread also execute with more times of iteration than that in the core workflow to obtain more accurate odometry results. The transformation from these modules are integrated. 

\section{Experiments}
\begin{figure}
  \centering
  \includegraphics[scale = 0.95]{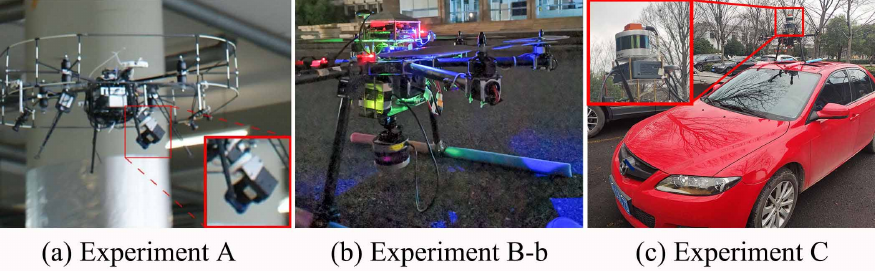}
  \caption{Sensor configurations in experiments. (a) A MAV with a spinning 2D lidar \cite{droeschel2016multilayered} in the experiment A. (b) A MAV with a 3D lidar and onboard computer in the outdoor test in the experiment B. (c) A 3D lidar and an IMU on passenger vehicle in the experiment C. }
  \label{fig:8}
\end{figure}

We conducted several experiments to evaluate the performance of our system from different aspects and compare it with two state-of-the-art methods according to the scenarios. The algorithm is implemented by C++ based on ROS (Robot Operating System) in Linux, and run on an Intel NUC computer with a 2.7 GHz i7-8559U CPU inside. We test data from two custom types of light-weigh sensors, a spinning 2D Hokuyo UTM-30LX-EW lidar in the experiment A, and a Velodyne VLP-16 3D lidar in the others. Fig. \ref{fig:8} shows the sensor configurations. The main parameters in our algorithm include the side length of the cell $a$ and the interval $r$ between the test locations. They were both set mainly according to the scale of scenarios. In detail, for the small indoor test in the experiment B(a), $a$ = 0.4 m and $r$ = 0.4/6 m. In the larger parking garage in the experiment A, $a$ = 1.5 m and $r$ = 0.25 m. In the outdoor tests, $a$ = 1.8 m and $r$ = 0.3 m. When compared with those feature-based methods, the resolution of the feature points in their map is set identical with the interval between our test locations. A video attachment presenting the experiment process can be found in website$\footnote{https://www.youtube.com/watch?v=2nRJThK0hCw}$ .

\subsection{Registering Sparse Point Cloud}
We use a data set collected by a spinning 2D lidar mounted on a micro aerial vehicle (MAV) \cite{droeschel2016multilayered} in a parking garage to demonstrate the robustness of our method when faced with sparse point cloud. The data set contains 200 frames of 3D data assembled from a 2D laser scan with the aid of visual odometry. The low-resolution point clouds are particularly sparse. Thus, the registration becomes challenging for ICP method as shown in Section \RNum{4}-C. The overall trajectory length is 73 m.

\begin{figure}
  \centering
  \includegraphics[scale = 0.95]{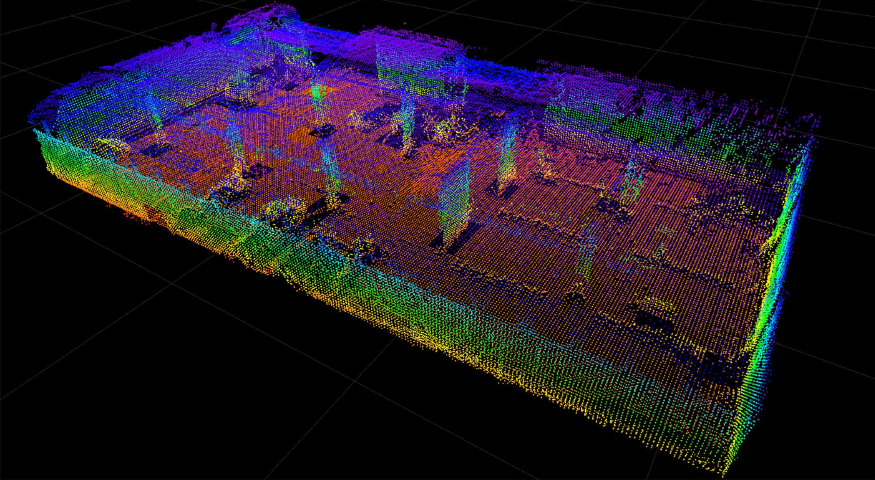}
  \caption{Map generated by GP-SLAM+ with spinning 2D lidar from the ``Parking garage'' data set. The map depicts rich details and recovers the walls without distortion. Points are colored according to height. }
  \label{fig:9}
  %\vspace{-0.4cm}
\end{figure}

The mapping result by our method is shown in Fig. \ref{fig:9}. The map recovers the structure of the garage and the walls show few distortion. The dense and uniformly distributed point-cloud-like map depicts rich details inside the building. By contrary, as shown in the Fig. 18(c) in work \cite{droeschel2016multilayered}, the GICP produces distort map even with graph optimization and registration with local dense map.

\subsection{Evaluation of the Core Workflow}
In the second part of experiments, we test the performance of the core workflow in our system without IMU. Here, we use one open-access method, A-LOAM$\footnote{https://github.com/HKUST-Aerial-Robotics/A-LOAM}$, as the benchmark, which is an advanced implementation of LOAM \cite{zhang2014loam}.

\subsubsection{Accuracy of State Estimation}
We evaluate the accuracy of state estimation with the ground-truth recorded by an Optitrack motion caption system. The range data were collected by a hand-held Velodyne VLP-16 lidar at a walking speed of 0.35 m/s in a room. The overall length of the trajectory is 53 m.

\begin{figure}
  \centering
  \includegraphics[scale = 0.43]{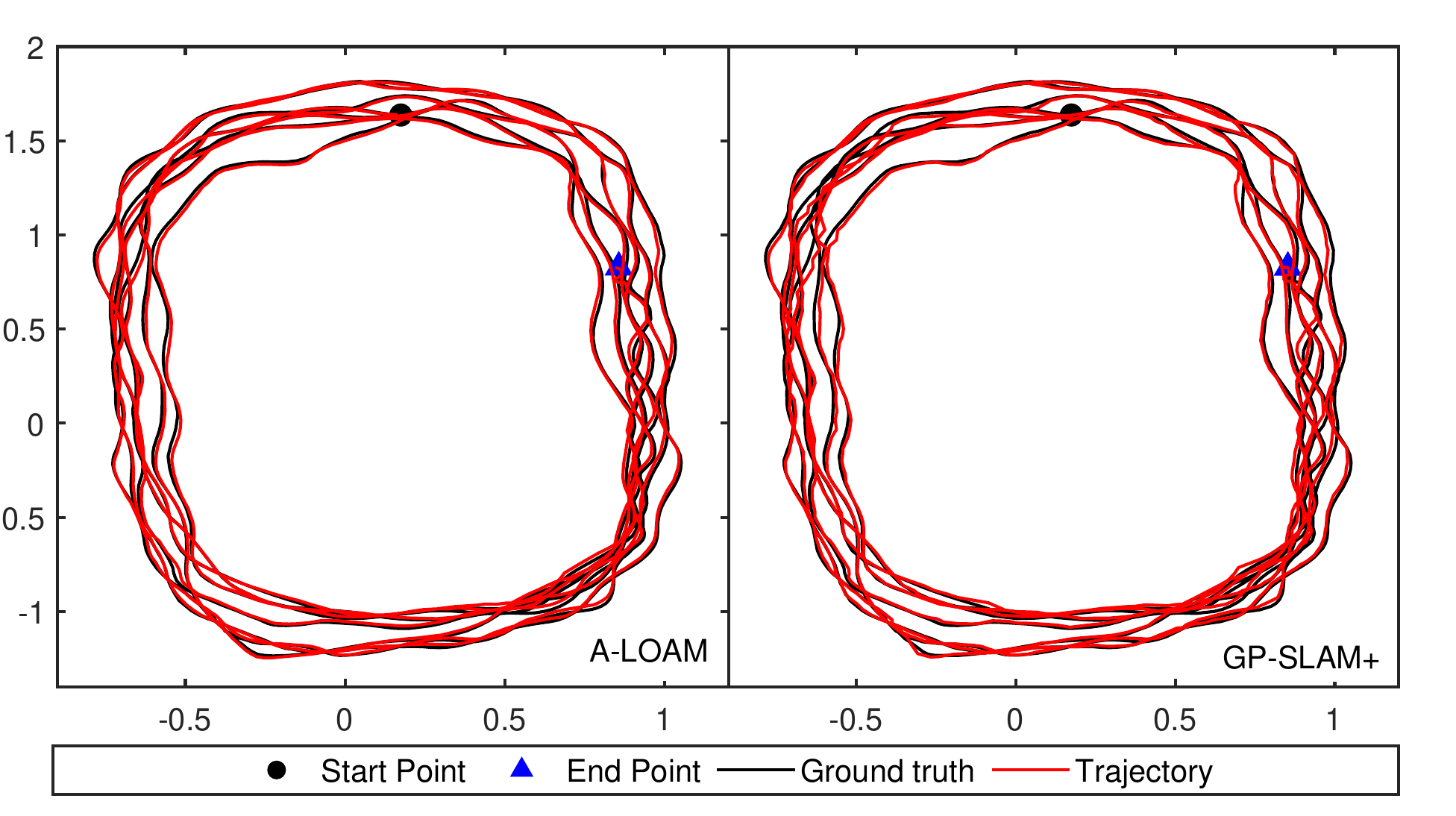}
  \caption{Overhead view of the trajectories produced by A-LOAM and GP-SLAM+ overlaid with ground truth in the indoor test.}
  \label{fig:10}

\end{figure}

The trajectory estimated by both methods are shown in Fig. \ref{fig:10}. Both GP-SLAM+ and A-LOAM yield relative precise pose estimation. For quantitative comparison, we align the trajectories with the ground-truth respectively, and calculate the average translation error. As reported in Table \ref{tab:1}, our method produces competing accuracy in state estimation comparable to A-LOAM.

\subsubsection{Quality of Mapping Result}
We mounted the sensor horizontally on a MAV to complete an outdoor mapping task. With the sensor suite shown in Fig. \ref{fig:8}-b, we finished this mapping task onboard. The plaza is surrounded by buildings, and the scale of it is 150$\times$120 m. During this experiment, the MAV took off from the north part, and finally landed on the south part after about one and a half circle. The length of the trajectory is about 200 m.

\begin{table}[]
	\centering
	\caption{Evaluation of State Estimation and Mapping}
	\label{tab:1}
	\begin{tabular}{@{}lll@{}}
		\toprule
		& A-LOAM & GP-SLAM+        \\ \midrule
		Avg. transl. error (m) & 0.0175 & {0.0156} \\
		MME                     & 1.5422 & {1.2014} \\ \bottomrule
	\end{tabular}
	%\vspace{-0.6cm}
\end{table}

\begin{figure}
	\centering
	\includegraphics[scale = 0.95]{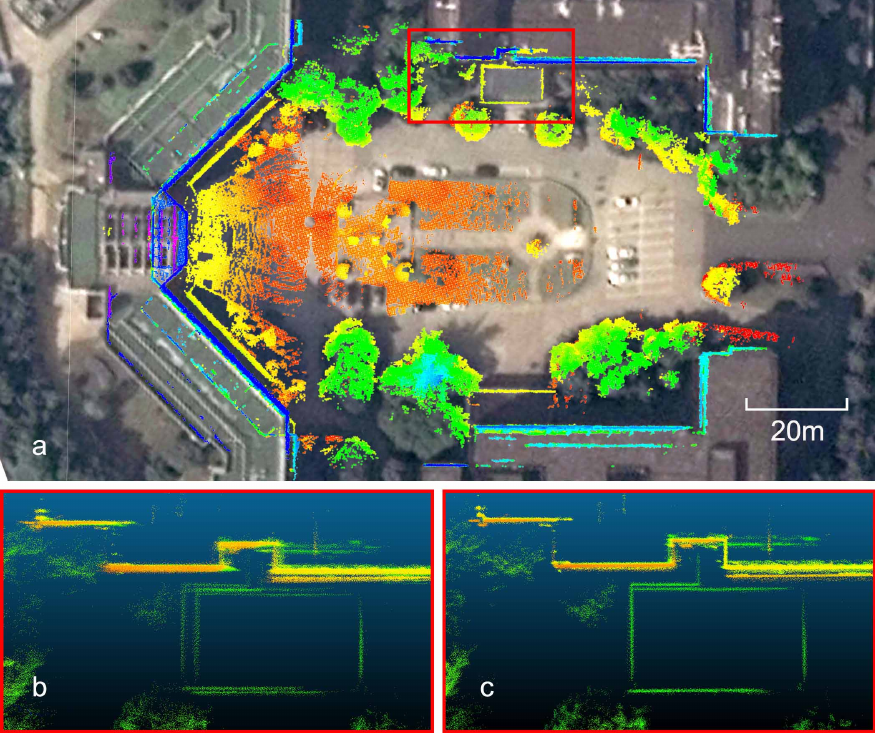}
	\caption{Qualitative analyze of the mapping result in the aerial test. Points are colored according to height. (a) Map produced by GP-SLAM+ overlaid on the satellite image. (b)(c) are the partial view of the aggregated raw point cloud by A-LOAM and GP-SLAM+ respectively. The map of A-LOAM contains multi-wall phenomenon.}
	\label{fig:11}
\end{figure}

The mapping result of the core workflow in GP-SLAM+ is presented in Fig. \ref{fig:1}. The map shows rich details including trees and sculptures. When overlaid on the satellite image (Fig. \ref{fig:11}), the map exhibits good alignment with it, and the maximal gap is less than 1 m measured manually. The map contains no multi-wall phenomenon, which demonstrates the accuracy of it. Furthermore, we check the aggregation of registered raw point cloud from both methods in the partial views in Fig. \ref{fig:11} b\&c. We use the mean map entropy (MME) as the criteria to evaluate the consistency of the registered raw data with the tool from \cite{razlaw2015evaluation}. The searching radius in the tool is set as 1.5 m in this outdoor scenarios. As listed in Table \ref{tab:1}, our method outputs smaller entropy, which means the point cloud registered by our method are sharper compared with that by A-LOAM. 

\begin{figure}
  \centering
  \includegraphics[scale = 0.95]{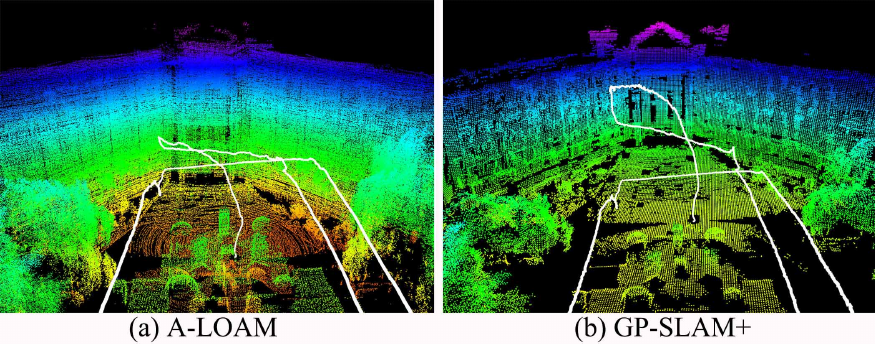}
  \caption{Comparison between A-LOAM and GP-SLAM+ in a degenerate scene. Points are colored according to height. White curve indicates the estimated trajectory of the MAV. (a) A-LOAM fails to track the motion due to lack of features and the map gets blur. (b) GP-SLAM+ produces consistent mapping result. }
  \label{fig:12}
  %\vspace{-0.6cm}
\end{figure}

 At the end of this task, the MAV climbed up and down to model a building. The structure of each floor on the wall are similar with less vertical features, and the amount of valid measurements becomes less at the top of the path. Therefore, it forms a degenerate scene for lidar-based method. As shown in Fig. \ref{fig:12}, A-LOAM fails to recover the motion and the map becomes blurred, while our method still yields reliable odometry and consistent map. The reason is that the feature extraction strategy prone to loss more valid information compared to our approach.

\subsubsection{Efficiency }
Considering only odometry can not yield consistent map, here we compare the efficiency of the scan-to-map registration process in both methods. During the aerial test, our core workflow completes both odometry and mapping in 73 ms for each frame, whereas the mapping thread in A-LOAM takes 224 ms per step. Notice that here the interval between range data is 100 ms. The mapping thread in A-LOAM will drop data automatically if it cannot process it in time. When this occurs, and this data will only be processed in their odometry thread. Our method processes all the 3842 frames of range data, while the mapping thread in A-LOAM processes 1774 frames of them. Thus, our method achieves better real-time performance.

To assess efficiency further, we break down the time consumption of both methods into four main modules including preprocessing, matching, alignment and map building. As listed in Table \ref{tab:3}, the preprocessing containing GP map reconstruction is the main computational burden of GP-SLAM+. Given $n_{cur}$ training points in $n_{cell}$ cells in the current frame, the computational time complexity of GP map reconstruction is ${O}\left({n}_{cur}^{3} / {n}_{cell}^{2}\right)$, where  $n_{cur}$ is typically one magnitude larger than $n_{cell}$. $n_{cur}$ is reduced by our principled down-sample filter. We utilize the evenly distributed property of samples in the matching and the map building processes so that they can be finished in  ${O}\left(n_{cur}\right)$ time. For those ICP-based methods, the main cost is the searching for closest points. Although this process is accelerated by Kd-tree, the building of this data structure still cost ${O}\left( {n}_{{map}} \log {n}_{{map}}\right)$ and the entire searching time is ${O}\left( {n}_{{cur}} \log {n}_{{map}}\right)$ \cite{bentley1975multidimensional}. Concerning the lidar-based SLAM system, to restrain the pose drift, the map is usually denser and $n_{map}$ can be one or more magnitude larger than $n_{cur}$. For instance, in this outdoor test, the scale of the average $n_{map}$ and $n_{cur}$ are $10^6$ and $10^4$ in A-LOAM. Therefore, our method employs a different strategy that focuses on the preprocessing step compare with ICP-based methods.

% \usepackage{multirow}
\begin{table}[]
	\centering
	\caption{Computation Time Break-down of Modules in Mapping}
	\label{tab:3}
	\begin{tabular}{@{}llll@{}}
		\toprule
		\multicolumn{2}{l}{Method}                                                                                 & A-LOAM & GP-SLAM+    \\ \midrule
		\multirow{5}{*}{\begin{tabular}[c]{@{}l@{}}Avg. time cost in a \\ mapping frame (ms)\end{tabular}} & Preprocessing & 5      & {58} \\
		& Matching    & 134    & {1}  \\
		& Alignment   & 36     & {13} \\
		& Map building  & 49     & {1}  \\
		& All         & 224    & {73} \\ \bottomrule
	\end{tabular}
	%\vspace{-0.4cm}
\end{table}

\begin{table}[]
	\centering
	\caption{Accuracy of Pose Estimation in Large-Scale Test}
	\label{tab:4}
	\begin{tabular}{@{Method}llll@{}}
		\toprule
		& \begin{tabular}[c]{@{}l@{}}LeGO-\\ LOAM\end{tabular} & \begin{tabular}[c]{@{}l@{}}GP-\\ SLAM+\end{tabular} & \begin{tabular}[c]{@{}l@{}}Full GP-\\ SLAM+\end{tabular} \\ \midrule
		Avg. transl. error in x-y (m) & 7.355                                                & 5.753                                               & 4.032                                                    \\
		Final elevation error (m)     & 42.136                                               & 5.561                                               & 0.178                                                    \\ \bottomrule
	\end{tabular}
	\vspace{-0.4cm}
\end{table}

\subsection{Evaluation of the Full System}
Finally, we test the full system in a large-scale task. The VLP-16 lidar together with an Xsens MTi-610 IMU was mounted upon a passenger vehicle. The ground-truth was provided by an RTK-GPS. The vehicle traveled 2.1 km in a campus at an average speed of 2.7 m/s. Since A-LOAM do not utilize IMU information, here we choose LeGO-LOAM \cite{shan2018lego} as the baseline to conduct a fair comparison. It performs higher efficiency compared with the original LOAM and is optimized for ground application, which also means it is not suitable for the aerial experiment B directly.

\begin{figure}
  \centering
  \includegraphics[scale = 1.00]{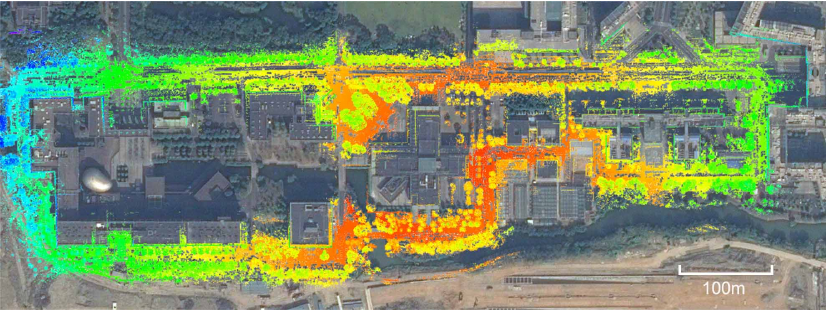}
  \caption{Mapping result of the full GP-SLAM+ in large-scale test overlaid on satellite image. The overall run is 2.1 km and the average vehicle speed is 2.7 m/s. Points are colored according to height. }
  \label{fig:13}
\end{figure}

\begin{figure}
  \centering
  \includegraphics[scale = 0.45]{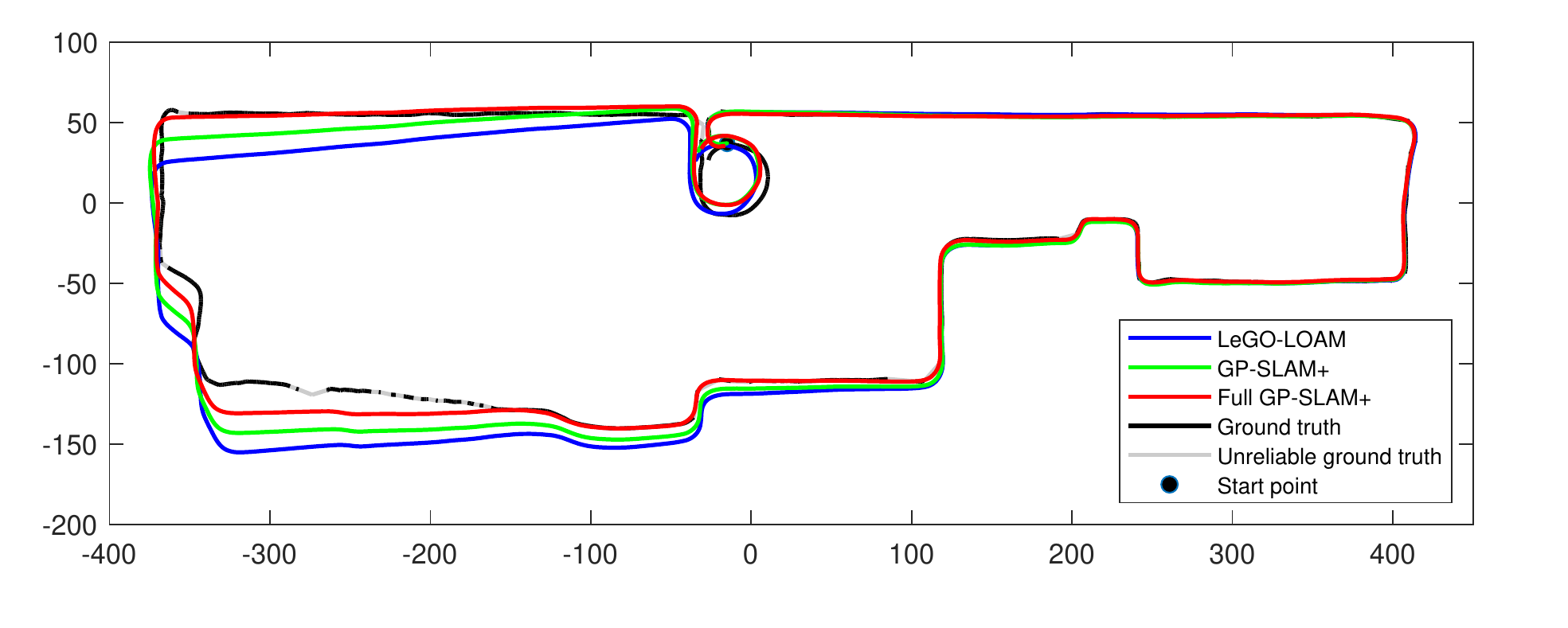}
  \caption{Overhead view of the trajectories in the large-scale test. It shows the trajectories from the core workflow (green) and refinement thread (red) in GP-SLAM+, LeGO-LOAM (blue) and ground truth (black).  }
  \label{fig:14}
  %\vspace{-0.6cm}
\end{figure}

This scenario includes urban and unstructured environment. We overlay the mapping result produced by our refinement thread with satellite image in Fig. \ref{fig:13}. Our method produces coherent map. To visualize the drift, we align the front 30\% part of the trajectories produced by two methods with the ground-truth respectively, and draw them in Fig. \ref{fig:14}. Due to occlusion from buildings or trees, the RTK-GPS signal is unavailable in the southwest part. In those areas, the accuracy can be demonstrated by the map and satellite image in Fig. \ref{fig:13}. For quantitatively evaluation, we align the entire trajectories with the ground truth respectively. Then, in Table \ref{tab:4}, we calculate the average translation error in x-y plane and the elevation difference when the vehicle returned to the start point. As it shows, the core workflow in our method accumulates less pose drift compared with LeGO-LOAM, and the refinement thread further enhances the performance especially in terms of the elevation error.

% \usepackage{multirow}
% \usepackage{booktabs}
%                                                                        & LeGO-LOAM & GP-SLAM+ & Full GP-SLAM+ \\ \midrule
%\begin{tabular}[c]{@{}l@{}}Avg. transl. \\ error in x-y(m)\end{tabular} & 7.355     & 5.753    & 4.032         \\
%\begin{tabular}[c]{@{}l@{}}Final elevation \\ error(m)\end{tabular}     & 42.136    & 5.561    & 0.178         \\ \bottomrule

\section{Discussion and Future Work}

The robustness of our method when registering sparse point cloud mainly derives from the GP map reconstruction, which devotes to model the local surfaces rather than focuses on points or features separately. The resulted maps show that the structure in areas that are sparsely covered by laser can be depicted clearly after our map building approach. The evenly distributed samples enable our core workflow to accomplish the scan-to-map registration in real-time. By contrast, the two baselines drops data in their mapping thread. Regarding the scan-to-map registration produces more precise estimation than the scan-to-scan one, we believe this efficiency to be one of the reasons that our method produced more accurate pose estimation. These two main advantages of our method, robustness with sparse point cloud and efficiency, was not obvious in small space (e.g., the room in the experiment B-a) but was significant in the outdoor tests. Although we use a filter in the GP map reconstruction process, the principle down-sampling process mainly skips the redundant points due to the sweeping mechanism.

Our work demonstrates the advantages of using spatial GP to model the structure for SLAM system. It also opens up the possibility to various kernel learning techniques for Gaussian process which have been studied in machine learning literature. We will further explore them to obtain better performance. For instance, the accuracy of the model can be refined if online learning of the kernel function is employed like \cite{plagemann2008learning}. Moving into Hilbert space is also promising as claimed in some works \cite{doi:10.1177/0278364916684382}. Besides, We will investigate the impact when denser range data, like those produced by 64-channel lidar in KITTI benchmark, is applied.

%\newpage

\bibliographystyle{IEEEtran}
\bibliography{mybib}

\end{document}